# Comment on "Clustering by fast search and find of density peaks"

Shuliang Wang[1,2], Dakui Wang[2], Caoyuan Li[1], Yan Li[1]

School of software, Beijing Institute of Technology, Beijing, China

International School of Software, Wuhan University, Wuhan, China

Email: slwang2011@bit.edu.cn

**Abstract.** In [1], a clustering algorithm was given to find the centers of clusters quickly. However, the accuracy of this algorithm heavily depend on the threshold value of $d_c$. Furthermore, [1] has not provided any efficient way to select the threshold value of $d_c$, that is, one can have to estimate the value of $d_c$ depend on one's subjective experience. In this paper, based on the data field [2], we propose a new way to automatically extract the threshold value of $d_c$ from the original dataset by using the potential entropy of data field. For any dataset to be clustered, the most reasonable value of $d_c$ can be objectively calculated from the dataset by using our proposed method. The same experiments in [1] are redone with our proposed method on the same experimental datasets used in [1], the results of which shows that the problem to calculate the threshold value of $d_c$ in [1] has been solved by using our method.

## 1 Principles

The algorithm in [1] utilized Gaussian function to calculate density, which is quite similar with the way that data field used to calculate the potential of every point. [2] proposed the method to calculate potential of every point using Gaussian function in data field. For a dataset $\{x_1, x_2, x_3, \cdots, x_n\}$, the equation to calculate the potential of every point is:

$$\varphi(x) = \sum_{i=1}^{n}\left(e^{-\left(\frac{\|x-x_i\|}{\sigma}\right)^2}\right) \qquad (1)$$

Equation (1) is very similar to the equation that is used to calculate density in [1]. In data field, data points with larger potential located in the dense region. As shown in Figure 1, (a) shows the distribution of potential of data field, dark areas have larger potential. This is same with the density distribution of original data as shown in (b). Therefore, potential of data field and the density of points in [1] have the same effect.

Based on above analysis, the threshold value "$d_c$" in [1] can be calculated in the same way which is used to optimize the impact factor σ in data field. According to data field, if the potential of data is all the same, then the uncertainty of data distribution is the largest; if potential is uneven distributed, then the uncertainty is the smallest and potential can reflect the distribution of data. The uncertainty of data is usually represented by entropy, so we can use the entropy to optimize impact factor σ.

For a dataset $\{x_1, x_2, x_3, \cdots, x_n\}$, the potential of every point is $\{\varphi_1, \varphi_2, \varphi_3, \cdots, \varphi_n\}$, the equation to calculate entropy $H$ in data field is:

$$H = -\sum_{i=1}^{n}\frac{\varphi_i}{Z}log\left(\frac{\varphi_i}{Z}\right) \qquad (2)$$

In Equation (2), $Z = \sum_{i=1}^{n} \varphi_i$ is a normalization factor. For dataset used in Figure 1, the change of entropy with different $\sigma$ is shown in Figure 2.

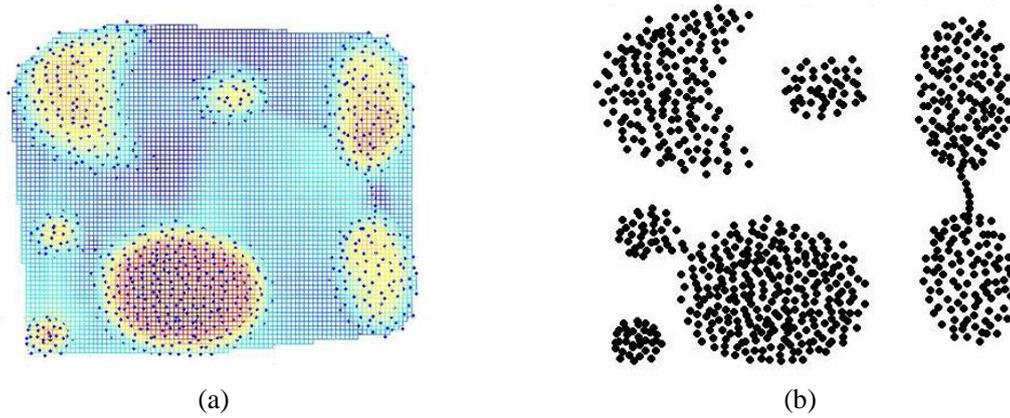

(a)            (b)

**Figure1. The potential distribution of data field**

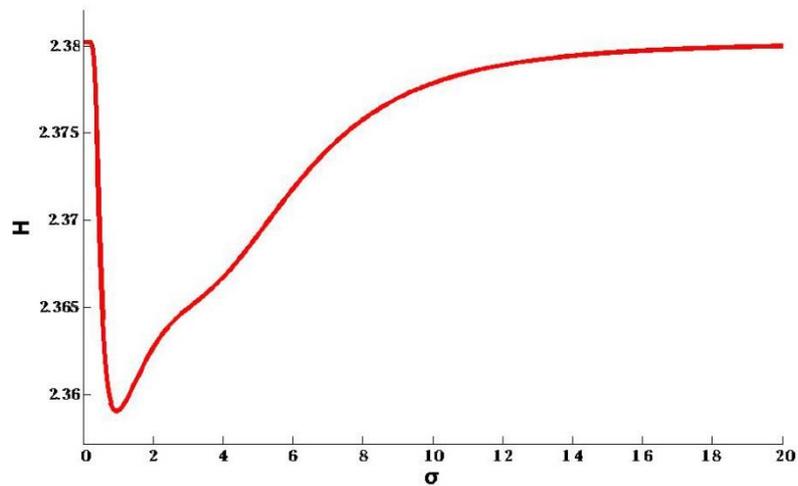

**Figure 2 The change of entropy with different σ**

In Figure 2, As $\sigma$ growing from 0 to ∞, the value of entropy decreases quickly at first, then increases slowly and finally maintains the same level. When the value of entropy is the smallest, the value of σ is 0.9531. With this $\sigma$, the distribution of potential in data field is shown as Figure 1, which reflects the distribution of data very well. As mentioned above, we should choose the value of $\sigma$ when the entropy is the smallest.

In data field, according to the 3B rule of Gaussian distribution [3], the influence radius is $\frac{3}{\sqrt{2}}\sigma$ for every point. As a clustering algorithm, one point can only affect the points located inside its influence radius, so we take $\frac{3}{\sqrt{2}}\sigma$ as a threshold value.

## 2 Experimental results and comparison

Our method is used to redo the experiments in [1] on the same experimental datasets used in [1]. Original datasets are shown in the second column of Figure 3:

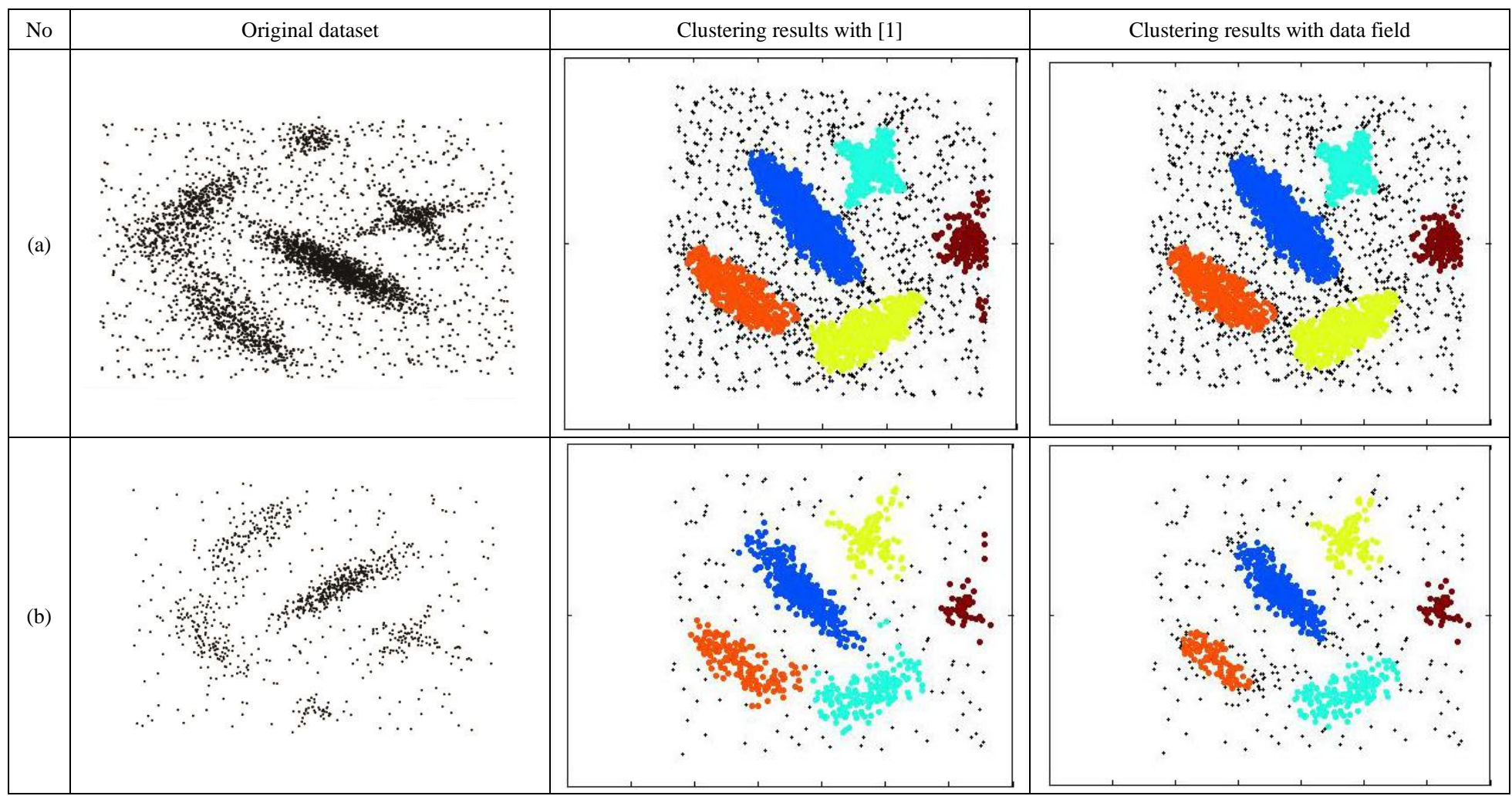

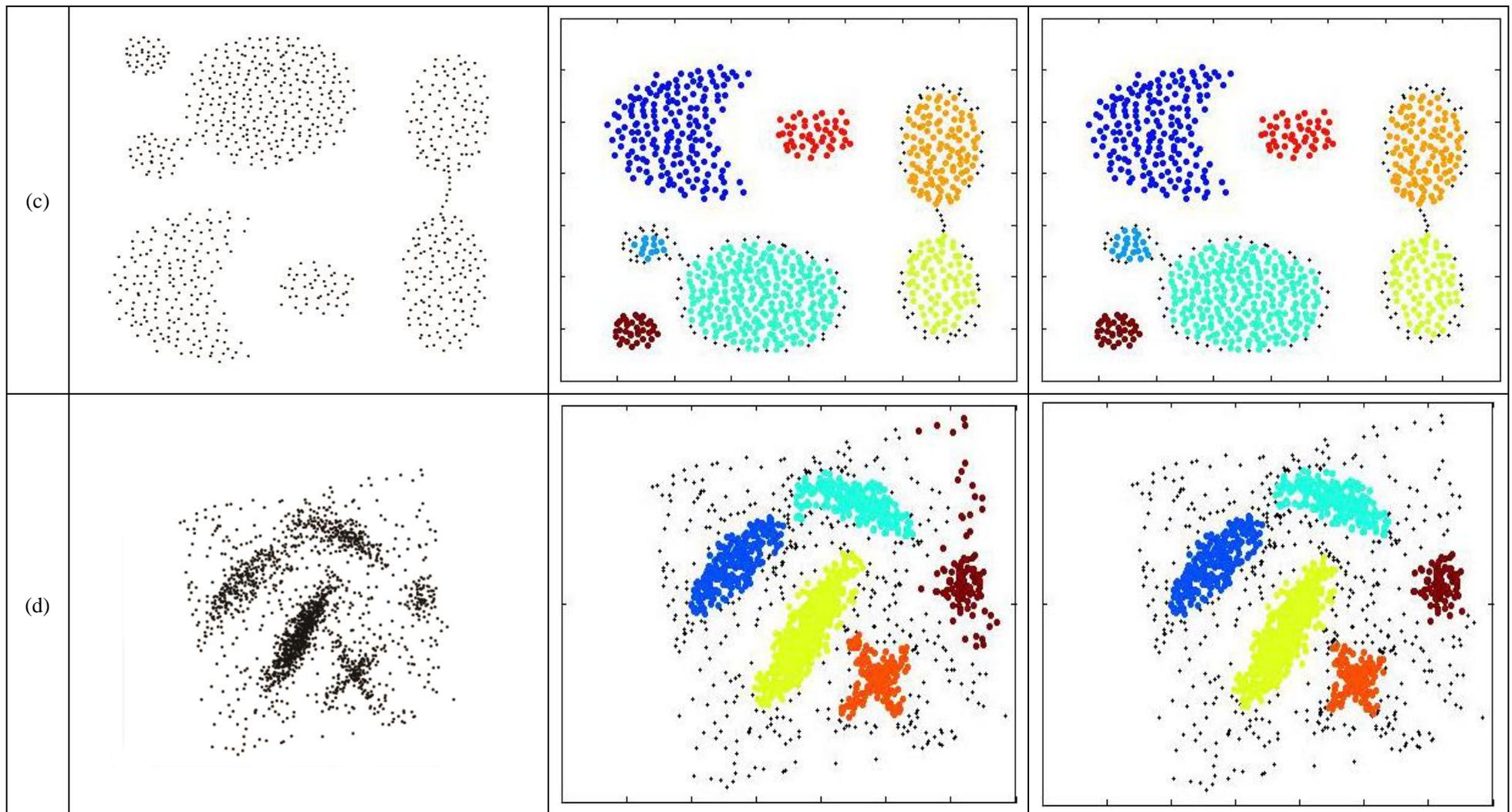

**Figure 4 The clustering result for each dataset with different threshold(continue)**

In Figure 3, there are four datasets named (a), (b), (c), (d). For the same dataset, we firstly use the estimated threshold value for clustering in [1], then we use the threshold calculated by data field for clustering again. The only difference between them is how the threshold is chosen.

In Figure 3, the third column is the clustering results with estimated threshold value in [1]. The fourth column is the clustering results with threshold that calculated by data field. According to Figure 3, the method to calculate the threshold with data field can remove the noisy points better than the estimated threshold value from [1]. The number of removed noisy points and threshold are shown in Table 1.

**Table 1 The number of removed noisy points and threshold**

|  | Algorithm in [1] | | Data field | |
| --- | --- | --- | --- | --- |
|  | Number of removed noisy points | Threshold | Number of removed noisy points | Threshold |
| Dataset (a) | 880 | 0.033039 | 904 | 0.040438 |
| Dataset (b) | 133 | 0.039533 | 240 | 0.060855 |
| Dataset (c) | 112 | 2.263846 | 100 | 2.021853 |
| Dataset (d) | 382 | 0.033000 | 450 | 0.045741 |

In Table 1, for dataset (a), (b), (d), which contains noisy points, our proposed method can remove more noisy points. For dataset (c), which does not contain any noisy points, data field method removed less non-noisy points. And in dataset (c), the threshold of original algorithm is similar to data field method. So the clustering results are almost the same.

With the comparison of Table 1 and Figure 3, it is clear that our proposed method can get better clustering result and higher clustering accuracy.

## 3 Conclusions

Our proposed method solved the problem in [1] that the threshold value cannot be calculated. Instead of estimating the value in [1], the threshold value of $d_c$ is calculate by using data field in this paper. With data field, one can get the best threshold value for different dataset automatically, which may improve the accuracy of the clustering algorithm in [1].

**Acknowledgement**
This work was supported by National Natural Science Fund of China (61472039, 61173061, and 71201120), Specialized Research Fund for the Doctoral Program of Higher Education (20121101110036), and the big data project of Yuan Provincial Energy Investment Group Co. LTD.

# References

[1] Alex Rodriguez, Alessandro Laio. Clustering by fast search and find of density peaks. Science, 27 JUNE 2014 • VOL 344 ISSUE 6191, 1492-1496.

[2] Shuliang Wang, Wenyan Gan, Deyi Li, Deren Li, Data Field for Hierarchical Clustering, International Journal of Data Warehousing and Mining, 7(4), 43-63

[3] Bárány, Imre; Vu, Van (2007), "Central limit theorems for Gaussian polytopes", Annals of Probability (Institute of Mathematical Statistics) 35 (4): 1593–1621